\documentclass[runningheads]{llncs}

\usepackage{eccv}

\usepackage{eccvabbrv}
\usepackage{indentfirst}
\usepackage{lineno}
\usepackage{subcaption}
\usepackage{arydshln}
\usepackage{multirow}
\usepackage{longtable}
\usepackage{color}
\usepackage{colortbl}
\usepackage{cite}
\usepackage{booktabs}
\usepackage{graphicx}
\usepackage{algorithm}
\usepackage{algorithmic}
\usepackage{stfloats}
\usepackage{amssymb,amsmath}
\usepackage{epstopdf}
\usepackage{rotating}
\usepackage{threeparttable}
\usepackage{bbding}
\usepackage{comment}

\usepackage[accsupp]{axessibility}  

\usepackage{hyperref}

\usepackage{orcidlink}

\begin{document}

\title{Frequency-Spatial Entanglement Learning for Camouflaged Object Detection} 

\titlerunning{FSEL}

\author{Yanguang Sun\inst{1} \and
Chunyan Xu\inst{1} \and
Jian Yang\inst{1}\and
Hanyu Xuan\inst{2} \textsuperscript{\Envelope} \and
Lei Luo\inst{1} \textsuperscript{\Envelope}}

\authorrunning{Y. Sun et al.}

\institute{PCA Lab, Nanjing University of Science and Technology, Nanjing, China \and
School of Big Data and Statistics, Anhui University, Heifei, China\\
\email{Sunyg@njust.edu.cn}}

\maketitle
\let\thefootnote\relax\footnotetext{\textsuperscript{\Envelope} Corresponding Author}

\begin{abstract}
Camouflaged object detection has attracted a lot of attention in computer vision. The main challenge lies in the high degree of similarity between camouflaged objects and their surroundings in the spatial domain, making identification difficult. Existing methods attempt to reduce the impact of pixel similarity by maximizing the distinguishing ability of spatial features with complicated design, but often ignore the sensitivity and locality of features in the spatial domain, leading to sub-optimal results. In this paper, we propose a new approach to address this issue by jointly exploring the representation in the frequency and spatial domains, introducing the Frequency-Spatial Entanglement Learning (FSEL) method. This method consists of a series of well-designed Entanglement Transformer Blocks (ETB) for representation learning, a Joint Domain Perception Module for semantic enhancement, and a Dual-domain Reverse Parser for feature integration in the frequency and spatial domains. Specifically, the ETB utilizes frequency self-attention to effectively characterize the relationship between different frequency bands, while the entanglement feed-forward network facilitates information interaction between features of different domains through entanglement learning. Our extensive experiments demonstrate the superiority of our FSEL over 21 state-of-the-art methods, through comprehensive quantitative and qualitative comparisons in three widely-used datasets. The source code is available at: \href{https://github.com/CSYSI/FSEL}{\color{blue}https://github.com/CSYSI/FSEL}.
\end{abstract}

\section{Introduction}
\label{sec:intro}
``Camouflage'' is a natural defense mechanism used by certain animals, such as chameleons, grasshoppers, and caterpillars, to blend into their surroundings and protect themselves. The study of camouflaged object detection (COD) focuses on identifying concealed targets in real-world situations. This research is crucial in developing robust visual perception models in computer vision. COD has a wide range of applications, including medical image analysis \cite{MIS,pranet}, species conservation \cite{SC}, and industrial defect detection \cite{cd}.

\begin{figure}[t]
	\centering\includegraphics[width=0.75\textwidth,height=3cm]{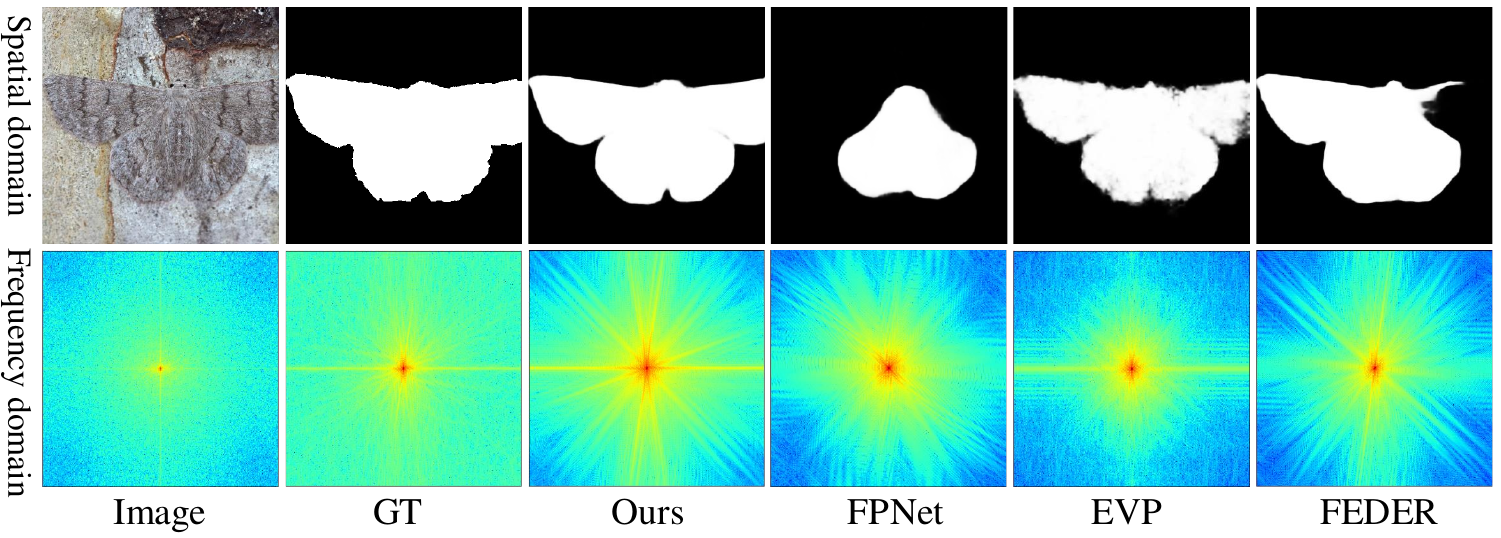}
	\caption{The visual comparison results of the proposed FSEL and current COD methods ($i.e.$, FPNet \cite{FPNet}, EVP \cite{EVP}, and FEDER \cite{FEDER}) in the spatial and frequency domain.}
	\label{Fig.1}
\end{figure}

In the early stages, some COD methods \cite{E_3,E_2,E_1} relied on manually crafted features to detect camouflaged objects. However, due to the extremely challenging appearance of these objects, the results obtained were often unsatisfactory. Later, with the advancement of deep learning and the availability of large-scale datasets \cite{CAMO,COD10K}, many COD methods \cite{JSOCOD,MGL,NC4K,SegMaR,ZoomNet} based on deep learning have been proposed. With a large amount of data available for training, these methods have the potential to automatically extract features and detect camouflaged objects, resulting in impressive performance. More recently, researchers have introduced various techniques for exploring useful information of input features from the spatial domain using boundary-guided \cite{BSANet,BGNet,FAPNet}, multi-scale strategy \cite{ZoomNet,SINet2,C2FNet}, uncertainty-aware \cite{JSOCOD,UGTR}, distraction mining \cite{PFNet}, and $etc$. 


We have observed that the COD methods \cite{COD10K,JSOCOD,MGL,BSANet,SegMaR} mentioned above primarily focus on single spatial features. While these spatial features are advantageous for COD tasks, they are often susceptible to interference from complex backgrounds. This vulnerability arises from their reliance on pixel-level information, with a primary emphasis on the local intensity and spatial position of individual pixels. Furthermore, spatial features possess local properties, meaning that pixels within a feature may only exhibit certain correlations with surrounding pixels. That being said, relying solely on spatial features can make it challenging to distinguish subtle variations within concealed objects and backgrounds. Therefore, it is crucial to find ways to overcome the limitations of spatial features to achieve accurate COD results. Recently, frequency features generated through Fourier Transform have been shown to have global characteristics and have been proven to be beneficial for understanding image contents \cite{FcaNet,FSP,FFC}. This can help break the bottleneck of spatial features.

Some recent COD methods \cite{FPNet,FDNet,EVP,FEDER,FA} have begun to incorporate frequency clues in their approach. These methods can be divided into two categories based on their objectives. The first category ($e.g.$, FDNet \cite{FDNet} and EVP \cite{EVP}) is to act directly on input images through different frequency transforms to extract frequency features, which are then combined with spatial features. However, camouflaged images often contain a lot of background noise, making the frequency features obtained from the image unreliable. When aggregated with spatial features, this may introduce some unnecessary background noises, resulting in under-segmented results (as depicted EVP \cite{EVP} in Fig. \ref{Fig.1}). The second category \cite{FPNet,FEDER,FA} focuses on initial features from the encoder. For example, Cong $et$ $al.$ \cite{FPNet} designed a frequency-perception module to improve the detection of camouflaged objects by utilizing both high-frequency features and low-frequency features. And He $et$ $al.$ \cite{FEDER} proposed frequency attention modules that obtain important parts of corresponding features by considering both high-frequency and low-frequency components. Although these methods have shown promising results, they only focus on high-frequency and low-frequency features, overlooking some information that falls between these two frequencies. This can be seen in FPNet \cite{FPNet} and FEDER \cite{FEDER} in Fig. \ref{Fig.1},  where significant information within the frequency domain may be missed.


Based on the above discussion, we propose a novel method called Frequency-Spatial Entanglement Learning for accurate camouflaged object detection. Our method combines global frequency features and local spatial features to optimize the initial input features and enhance their discriminative ability.  Specifically, we first establish a Frequency Self-attention to obtain discriminative global frequency features, which models the correlation between each frequency band and learns the dependency relationships between different frequencies in input bands. Moreover, we introduce entanglement learning between the frequency and spatial features in the Entanglement Transformer Block, allowing them to mutually learn and collaborate for optimization. Furthermore, we extend the applicability of global frequency features by utilizing the Joint Domain Perception Module and the Dual-domain Reverse Parser to optimize the input features and generate powerful representations that incorporate both frequency and spatial information. Extensive experiments on three widely-used benchmark datasets ($i.e.,$ CAMO \cite{CAMO}, COD10K \cite{COD10K}, and NC4K \cite{NC4K}) demonstrate that FSEL consistently outperforms 21 state-of-the-art COD methods across different backbones.

The main contributions can be summarized as follows:

(1) We propose a Frequency-Spatial Entanglement Learning (FSEL) framework that utilizes both global frequency and local spatial features to enhance the detection of camouflaged objects.
	
(2) To improve the representation capability of frequency and spatial features, we have designed an Entanglement Transformer Block (ETB). This block allows for entanglement learning of frequency-spatial features, resulting in a more comprehensive understanding of the data.

(3) To reduce the sensitivity and locality limitation of spatial features, we have incorporated frequency domain transformations into both the Joint Domain Perception Module (JDPM) and the Dual-domain Reverse Parser (DRP).

\section{Related work}
\textbf{Camouflaged Object Detection.} Recently, with the public availability of datasets ($i.e.$, CAMO \cite{CAMO}, COD10K \cite{COD10K}, and NC4K \cite{NC4K}), deep learning-based COD methods have started to surface in large numbers, which can be broadly categorized into, including multi-scale strategies \cite{TINet,ZoomNet,C2FNet}, edge-guidance \cite{BSANet,BGNet,FEDER,FAPNet}, uncertainty-aware \cite{UGTR,JSOCOD}, multi-graph learning \cite{MGL}, iterative manner \cite{SegMaR,HitNet,PreyNet1}, transformer \cite{FSNet,EVP}, and so on. Besides, other COD methods \cite{EVP,FPNet,FA} considered frequency clues to help with reasoning camouflaged objects. Particularly, Zhong $et$ $al.$ \cite{FDNet} processed directly the camouflaged image through discrete cosine transform to obtain frequency information. He $et$ $al$. \cite{FEDER} proposed frequency attention modules to filter out the noteworthy parts of corresponding features. After that, Cong $et$ $al.$ \cite{FPNet} designed a frequency-perception module by learning different frequency features to achieve coarse localization of camouflaged objects. However, these methods often focus on the high- and low-frequency information, ignoring the relationship between all bands in the frequency domain. Therefore, we conduct frequency analysis on different spectral features to achieve all frequency band interactions and importance allocation. Furthermore, we perform entanglement learning on global frequency and local spatial features, which is beneficial for obtaining powerful representations.
\begin{figure}[t]
	\centering\includegraphics[width=0.9\textwidth,height=6cm]{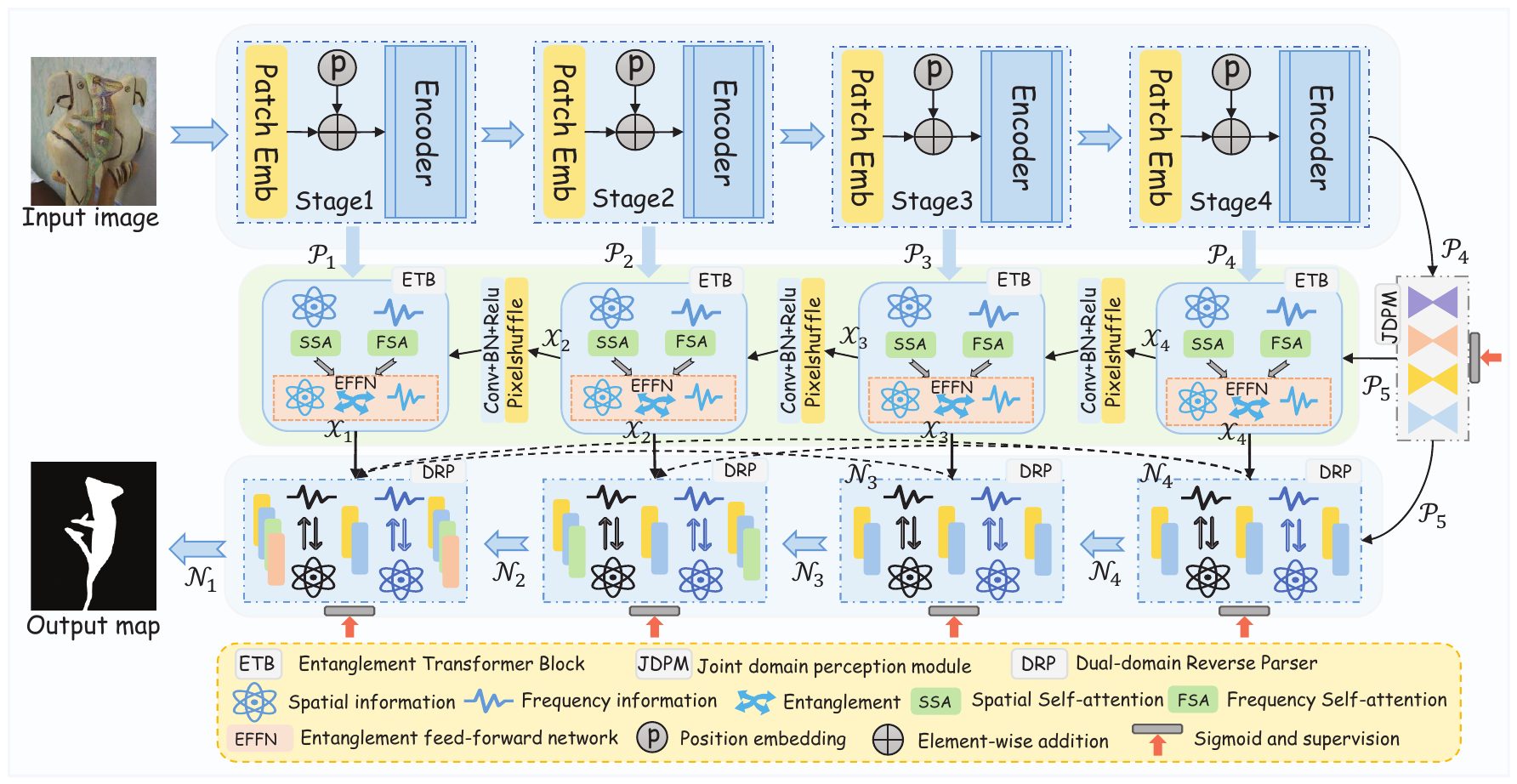}
        \caption{Overview of the proposed FSEL model framework for camouflaged object detection. The proposed FSEL method generates predicted results through a Joint Domain Perception Module (JDPM), a series of stacked Entanglement Transformer Block (ETB), and a Dual-domain Reverse Parser (DRP).
        }
	\label{Fig.2}
\end{figure}

\textbf{Vision Transformer.}
Transformer \cite{T} utilized self-attention to model global semantic and long-range dependencies, which helps to understand the correlation between different regions in an image, and therefore it has been widely used in some computer vision tasks, including object detection \cite{OB,DGT}, image classification \cite{BiFormer,PVT}, semantic segmentation \cite{HGFormer,Oneformer,co}, $etc$. For example, Yuan $et$ $al.$ \cite{ViT} obtained long-range relationships from the sequence of image patches to perform the image classification. Next, Liu $et$ $al.$ \cite{Swin} split input maps into non-overlapping local windows, and then transferred the information through shift operations between the windows to improve the efficiency of the model. In addition, other transformer models have been successful in computer visions, such as Restormer \cite{Restormer}, CrossFormer \cite{Cvt}, EfficientViT \cite{Evit}, MPFormer \cite{MPF}, and among others. Unlike these methods, which always model relationships based on the spatial domain, we transform spatial features into the frequency domain and combine them to perform dual-domain feature optimization.

\textbf{Frequency Learning.}
The frequency domain is very important for signal analysis, and recently it has been gradually applied in computer vision tasks. Particularly, Qin $et$ $al.$ \cite{FcaNet} assumed channel attention as a compression problem and introduced frequency transformation in the channel attention. Yun $et$ $al.$ \cite{SPANet} handled the balancing problem of different frequency components of visual features. Wang $et$ $al.$ \cite{FSP} proposed a frequency shortcut perspective in image classification. In addition, some frequency domain-based methods \cite{DFF,EVP,FPNet,FEDER,SPML,FFC} have achieved great performance. In this paper, we extend global frequency features to different applications, involving multi-receptive fields perception, transformer, and reverse attention.

\section{Method}

\subsection{Framework Architecture}\label{S2}
Camouflaged objects exhibit a high level of visual similarity to their backgrounds, achieved through adaptive changes in color, texture, and shape. This creates challenges in distinguishing between object and background pixels in the spatial domain. Additionally, the locality of features in the spatial domain is limited in understanding camouflaged objects. To address this issue, we have implemented several strategies: 1) We have expanded beyond the spatial domain and utilized Fourier transformation to map features to the frequency domain, allowing for a more global perspective; 2) We have analyzed the relationships between all frequency bands to combine global frequency features with local spatial features; 3) We have extended frequency features to multiple components to fully utilize the global understanding of the object.

The complete architecture of our FSEL model is shown in Fig. \ref{Fig.2}. Given an input image ${I_c}\in\mathbb{R}^{H\times W\times 3}$, we first use the basic encoder ($i.e.$, PVTv2 \cite{PVT}/ ResNet \cite{ResNet}/ Res2Net \cite{R2Net}) to extract initial feature $\mathcal{P}$$=$$\left \{\mathcal{P} _{i} \right \} _{i=1}^{4}$ with the resolution of $\frac{W}{2^{i+1}}$$\times$$\frac{H}{2^{i+1}}$. Then the JDPM (Sec. \ref{S3}) captures a higher-level semantic feature $\mathcal{P}_5$ to guide location by integrating multi-receptive field information from frequency-spatial domains. After that, the ETB (Sec. \ref{S4}) models the cross-long-range relationships and performs entanglement learning on the frequency and spatial domains from initial features to generate discriminative feature $\mathcal{X}$$=$$\left \{ \mathcal{X}_{i} \right \} _{i=1}^{4}$. To ensure the quality of predicted map $\mathcal{N}$$=$$\left \{ \mathcal{N}_{i} \right \} _{i=1}^{4}$, we design a DRP (Sec. \ref{S5}) to aggregate feature flows through auxiliary optimization in the frequency and spatial domains. 

\subsection{Joint Domain Perception Module}
\label{S3}

Multi-scale information is beneficial for contextual understanding in different regions. We observe that these methods \cite{ASPP, DenseASPP, RFB} often generate multi-scale features through different convolutions with multiple receptive fields in the spatial domain. However, the receptive field of convolution operations in the spatial domain is limited, and in the process of data processing, tiny fluctuations may be overlooked, resulting in sub-optimized outcomes.

\begin{figure}[t]
	\centering\includegraphics[width=0.78\textwidth,height=3.2cm]{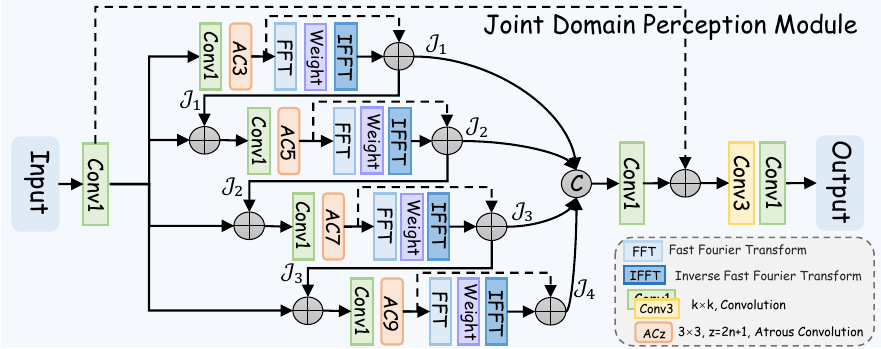}
	\caption{Details of the joint domain perception module.}
	\label{Fig.3}
\end{figure}
Therefore, we propose a Joint Domain Perception Module (JDPM) that reconstructs multi-receptive field information by introducing frequency transformation in multi-scale features. As depicted in Fig. \ref{Fig.3}, our JDRM uses the hierarchical structure to extract frequency-spatial information of different receptive fields. Technically, we use feature $\mathcal{P}_4$ as input and first reduce its channel numbers using 1$\times$1 convolution ($C_1$), $i.e.$, $\mathcal{P}_{4}^{128}=C_{1}\mathcal{P}_{4}$. Then, we construct a set of 3$\times$3 atrous convolutions ($\mathcal{AC}_{z}$) with filling rate $z$ to capture local multi-scale spatial feature $\left \{ \mathcal{J}_{n}^{s} \right \} _{n=1}^{4}$, $i.e.$, $\mathcal{J}_{n}^{s}=C_1\mathcal{AC}_{z}(\mathcal{P}_{4}^{128}+\mathcal{J}_{n-1})$, where $z=2n+1$ and $n-1\ge 1$. Next, we transform local spatial features into the frequency domain using the Fast Fourier Transform ($fft(\cdot)$) and perform redundancy filtering. We then use the Inverse Fast Fourier Transform ($ifft(\cdot)$) and the modulus of complex features to obtain global frequency features $\left \{ \mathcal{J}_{n}^{f} \right \} _{n=1}^{4}$, which is defined as:
\begin{equation}
	\begin{split}
		& \mathcal{J}_{n}^{f}=\Phi \left \|ifft(\sigma(fft(\mathcal{J}_{n}^{s}))*fft(\mathcal{J}_{n}^{s})) \right \|, n=1,2,3,4, \\
		& fft(u,v)=\sum_{x=0}^{W-1}\sum_{y=0}^{H-1}\mathcal{J}_{n}^{s}(x,y)e^{-2\pi i(\tfrac{ux}{W}+\tfrac{vy}{H})},\\ 
            & ifft(x,y)=\frac{1}{WH} \sum_{u=0}^{W-1}\sum_{v=0}^{H-1}fft(u,v)e^{2\pi i(\tfrac{ux}{W}+\tfrac{vy}{H})}, \\
	\end{split} 
\end{equation}
where $\Phi \left \| \cdot \right \|$ and ``$*$'' denote the modulus operation and the element-wise multiplication. $\sigma(\cdot)$ presents a set of weight coefficients, which sequentially contains a convolution, a batch normalization, a ReLU, a convolution, and a sigmoid function. $(u,v)$ and $(x,y)$ denote frequency domain coordinates and spatial domain coordinates, $i$ represents the imaginary part. After that, we aggregate global frequency features with local spatial features to generate intermediate multi-scale features $\left \{ \mathcal{J}_{n} \right \} _{n=1}^{4}$, that is, $\mathcal{J}_{n}=J_{n}^{s}+ \mathcal{J}_{n}^{f}, n=1,2,3,4$. 

Finally, we concatenate all multi-scale features and introduce residual connections to generate a coarse feature map $\mathcal{P}_5$ with 1-channel through a 3 $\times$ 3 and a 1 $\times$ 1 convolutions, which can be expressed as:
\begin{equation}
	\begin{split}
			& \mathcal{P}_5=C_3C_1(C_1Cat(\mathcal{J}_1,\mathcal{J}_2,\mathcal{J}_3,\mathcal{J}_4)+\mathcal{P}_{4}^{128}), 
	\end{split}  
\end{equation}
\label{E2}
where $C_k$ presents $k\times k$ convolution. $Cat(\cdot,\cdot,\cdot,\cdot)$ and ``+'' denote concatenation and element-wise addition.

\begin{figure}[t]
	\centering\includegraphics[width=0.85\textwidth,height=4.5cm]{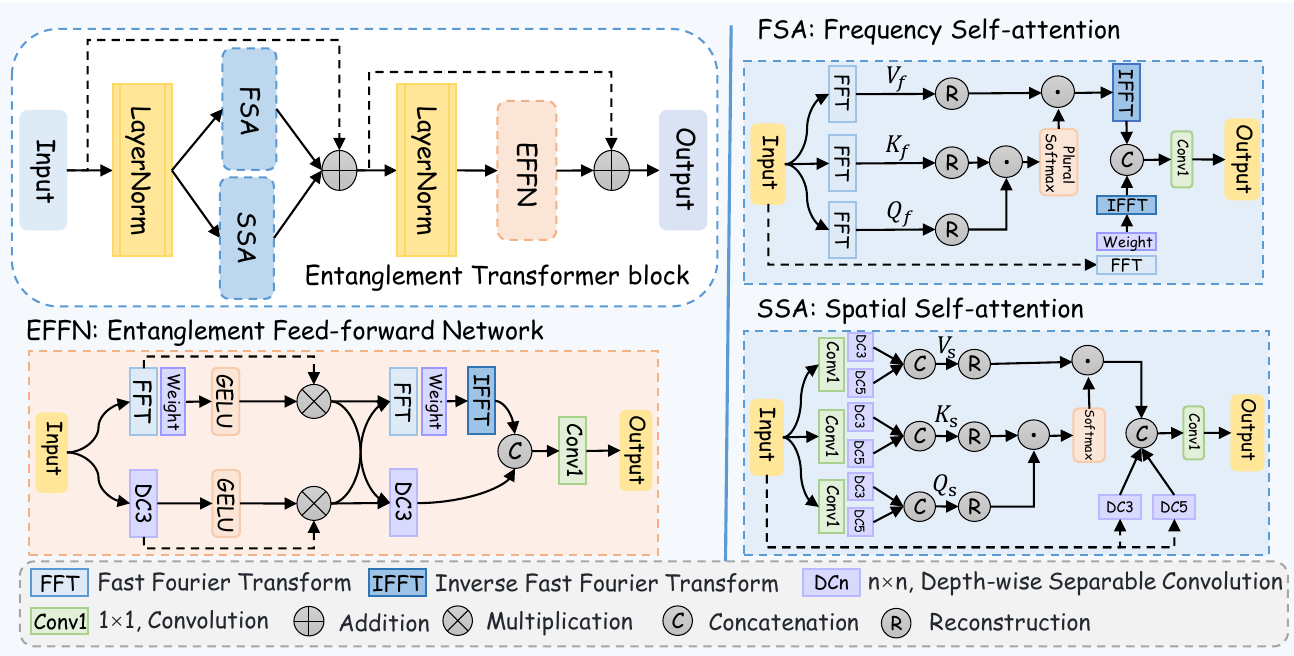}
	\caption{Details of the entanglement transformer block.}
	\label{Fig.4}
\end{figure}

\subsection{Entanglement Transformer Block}\label{S4}
Unlike previous methods \cite{FPNet,Swin,FSPNet}, which only model long-range dependencies based on local features in the spatial domain, our ETB incorporates different relationships from the frequency and spatial domains. In addition, we propose entanglement learning for different domain features in the ETB, allowing for the integration of information such as color, texture, edge, spectral, amplitude, and energy. This approach is beneficial for learning discriminative representations by considering various types of information. As depicted in Fig. \ref{Fig.4}, our ETB consists of three key components: frequency self-attention (FSA), spatial self-attention (SSA), and entanglement feed-forward network (EFFN).

\textbf{Frequency self-attention.}
To better analyze frequency signals, we introduce a self-attention structure, which allows the model to obtain the relationships and interactions among different frequency bands, learn importance weights, and perform adaptive fusion. Technically, the feature $\mathcal{P}_{\psi}^{128}$ with 128-channel is taken as an input, which comes from the initial features at the same level and the optimized high-level features (as depicted in Fig. \ref{Fig.2}), and a layer normalization to generate feature $\hat{\mathcal{P}}$$(\hat{\mathcal{P}}$$=$$\mathcal{LN}(\mathcal{P}_{\psi}^{128}))$. After that, we use a Fast Fourier Transform to obtain $query$ $Q_f$$=$$fft^{Q}(\hat{\mathcal{P}})$, $key$ $K_f$$=$$fft^{K}(\hat{\mathcal{P}})$, $value$ $V_f$$=$$fft^{V}(\hat{\mathcal{P}})$ in the frequency domain, where $fft^{(\cdot)}(\cdot)$ denotes the Fast Fourier Transform, and then we reshape $query$ ($\widetilde{Q}_f$$\in$$\mathbb{R}^{C\times HW}$) and $key$ ($\widetilde{K}_f$$\in$$\mathbb{R}^{HW\times C}$) projections that their dot-product generates transpose-attention map ($\Lambda_f$, $\Lambda_f$$=$$\widetilde{Q}_f\odot\widetilde{K}_f$), where $\odot$ presents the matrix multiplication. Different from these models \cite{FSPNet,HitNet} that directly utilize the Softmax function to activate the attention map with the real type, the frequency attention map ($\Lambda_f$) is a complex type that cannot be activated directly. Therefore, we extract the real part ($\Lambda _{f}^{re}$, $\Lambda _{f}^{re}$$=$$(\Lambda _{f}+conj(\Lambda _{f}))/2$) and imaginary part ($\Lambda _{f}^{im}$, $\Lambda _{f}^{im}$$=$$(\Lambda _{f}-conj(\Lambda _{f}))/2i$), where $conj(\cdot)$ denotes conjugate complex number, and then activate and merge the real and imaginary parts to obtain the activated attention maps ($a\Lambda _{f}$), as follows:
\begin{equation}
\begin{split}
	&a\Lambda _{f}=\Theta (Sof(\Lambda _{f}^{re}),Sof(\Lambda _{f}^{im})),\\
\end{split}
\end{equation}
where $\Theta(\cdot,\cdot)$ presents a combination function that combines the imaginary and real parts into a complex number. $Sof(\cdot)$ denotes a Softmax function. Subsequently, we use the attention map $a\Lambda _{f}$ to optimize the weights on the frequency feature $V_f$ and then use the Inverse Fast Fourier Transform ($ifft(\cdot)$) to convert it to an original domain and employ the modulus operation to obtain the frequency attention feature. In addition, we introduce a frequency residual connection to increase frequency information ($\hat{\mathcal{P}}_{f}^{r}$, $\hat{\mathcal{P}}_{f}^{r}$=$\Phi ||ifft(\sigma(fft(\hat{\mathcal{P}}))*fft(\hat{\mathcal{P}}))||$), and finally fuse features to produce the frequency feature $\mathcal{X}_f^{1}$, which is formulated as:
\begin{equation}
\begin{split}
	&\mathcal{X}_f^{1}=C_1Cat(\Phi ||ifft(a\Lambda _{f}\odot\widetilde{V}_f)||,\hat{\mathcal{P}}_{f}^{r}),\\
\end{split}
\label{E4}
\end{equation}
where $Cat(\cdot,\cdot)$ and $\odot$ are concatenation and matrix multiplication. $\Phi \left \| \cdot \right \|$ presents the modulus operation. $\widetilde{V}_f$ is the reshaped $V_f$.

\textbf{Spatial self-attention.}
Considering the unfixed size of camouflaged objects, we embed abundant contextual information into spatial self-attention. As shown in the bottom right of Fig. \ref{Fig.4}, similar to the FSA, we take the feature $\hat{\mathcal{P}}$ as the input and encode the position information using a 1$\times$1 convolution ($C_1$), and then we obtain the $query$ $Q_s$, $key$ $K_s$, and $value$ $V_s$ required by the self-attention by utilizing two depth-wise separable convolution with 3$\times$3 ($\mathcal{DC}_3$) and 5$\times$5 ($\mathcal{DC}_5$). After that, we generate the attention map ($a\Lambda _{s}$, $a\Lambda _{s}$$=$ $Sof$($\widetilde{Q}_s \odot\widetilde{K}_s$) ) through the reconstructed $\widetilde{Q}_s$ and $\widetilde{K}_s$ and activate it using Softmax function. Subsequently, the activated attention map $a\Lambda _{s}$ is used to correct the weights of $V_s$. Besides, to increase the spatial local information ($\hat{\mathcal{P}}_{s}^{r}$, $\hat{\mathcal{P}}_{s}^{r}$$=$$Cat(\mathcal{DC}_3\hat{\mathcal{P}}, \mathcal{DC}_5\hat{\mathcal{P}})$), we perform a residual connection to generate spatial feature $\mathcal{X}_s^{r}$, as shown in:
\begin{equation}
\begin{split}
	&\mathcal{X}_s^{r} = C_1Cat(a\Lambda _{s}\odot\widetilde{V}_s,\hat{\mathcal{P}}_{s}^{r}),\\
\end{split}
\end{equation}
where $C_1$, $Cat(\cdot,\cdot)$, and $\odot$ are the same as in Eq. {\color{red}(\ref{E4})}. $\widetilde{V}_s$ is the reshaped $V_s$.

\textbf{Entanglement feed-forward network.}
Frequency and spatial features usually contain different information. The frequency domain focuses on the global energy distribution and variation of signals, while spatial information acts on local pixel-level details and spatial structures, all of which are crucial for comprehending camouflaged objects. In our EFFN, these features are considered as two kinds of states that can perform entanglement learning to obtain more robust and powerful representations during the entanglement process.

Specifically, we first entangle the global frequency feature $\mathcal{X}_f^{1}$ and the local spatial feature $\mathcal{X}_s^{1}$ to adapt them to each other, followed by the residual connection to acquire the comprehensive feature $\mathcal{X}_c^{1}$, that is, $\mathcal{X}_c^{1}$$=$$\mathcal{X}_s^{1}$$+$$\mathcal{X}_f^{1}$$+$$\mathcal{P}_{\psi}^{128}$, which performs the layer normalization to improve the stability, and then the normalized feature $\hat{\mathcal{X}_c^{1}}$ ($\hat{\mathcal{X}_c^{1}}$=$\mathcal{LN}(\mathcal{X}_c^{1})$) is subjected to non-linearity entanglement learning in the EFFN. Technically, the EFFN consists of two phases, the first stage projects the feature $\hat{\mathcal{X}_c^{1}}$ to the frequency and spatial domains, and utilizes the GELU function for nonlinear activation and a gate mechanism to obtain global frequency feature $\hat{\mathcal{X}}_f^{2}$ and local spatial feature $\hat{\mathcal{X}}_s^{2}$, which can be written as follows:
\begin{equation}
\begin{split}
	&\hat{\mathcal{X}}_f^{2}= GE(\Phi ||\sigma(fft(\hat{\mathcal{X}_c^{1}}))*fft(\hat{\mathcal{X}_c^{1}})||)*\Phi||\sigma(fft(\hat{\mathcal{X}_c^{1}}))*fft(\hat{\mathcal{X}_c^{1}})||,\\
        &\hat{\mathcal{X}}_s^{2} = GE(\mathcal{DC}_3\hat{\mathcal{X}_c^{1}})*\mathcal{DC}_3\hat{\mathcal{X}_c^{1}},
\end{split}
\end{equation}
where $GE(\cdot)$ denotes the GELU function. Subsequently, in the second stage, the frequency and spatial features from the first stage are again entangled by interacting with each other by transferring information from different domains, and the entangled frequency-spatial features are optimized independently. They are then aggregated and reduced channels to generate comprehensive feature $\hat{\mathcal{X}}_c^{3}$, which can be formulated as follows:
\begin{equation}
\begin{split}
	&\hat{\mathcal{X}}_c^{3}=C_1Cat(\hat{\mathcal{X}}_{f}^{3},\hat{\mathcal{X}}_{s}^{3})+\mathcal{X}_c^{1},\\
        &\hat{\mathcal{X}}_{f}^{3}=\Phi ||ifft(\sigma(fft(Cat(\hat{\mathcal{X}}_{f}^{2},\hat{\mathcal{X}}_{s}^{2})))*fft(Cat(\hat{\mathcal{X}}_{f}^{2},\hat{\mathcal{X}}_{s}^{2})))||,\\
        &\hat{\mathcal{X}}_{s}^{3} =\mathcal{DC}_3Cat(\hat{\mathcal{X}}_{f}^{2},\hat{\mathcal{X}}_{s}^{2}), 
\end{split}
\end{equation}
where $C_1$, $Cat(\cdot,\cdot)$ and $\Phi||$$\cdot$$||$ are the same as in Eq. {\color{red}(\ref{E4})}. $fft(\cdot)$ and $ifft(\cdot)$ present the Fast Fourier Transform and the Inverse Fast Fourier Transform. $\mathcal{DC}_3$ denotes the depth-wise separable convolution with 3$\times$3 kernel. Finally, we introduce residual connections to obtain the final feature $\mathcal{X}$ with 128-channel in the ETB, $i.e.$, $\mathcal{X}$=$C_{1}Cat(\hat{\mathcal{X}}_c^{3},\mathcal{P}_{\psi}^{128})+\mathcal{P}_{\psi}^{128}$. Through multiple aggregation interactions, global frequency and local spatial features interact and depend on each other, leading to the entanglement of features from different states, forming rich and comprehensive representations.

\begin{figure}[t]
	\centering\includegraphics[width=0.80\textwidth,height=3.2cm]{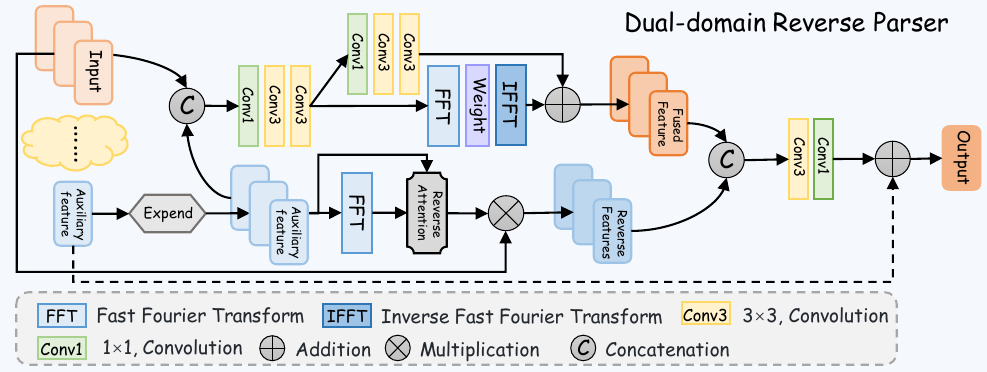}
	\caption{Details of the dual-domain reverse parser.}
	\label{Fig.5}	
\end{figure}

\subsection{Dual-domain Reverse Parser}\label{S5}
Different from these methods \cite{FSPNet,SINet2,BSANet} that integrate multi-level features based on the spatial domain, we propose the dual-domain reverse parser (DRP), which optimizes and aggregates diverse information from multi-level feature $\mathcal{X}$ in both frequency and spatial domains. As depicted in Fig. \ref{Fig.2}, we first take the feature $\mathcal{X}_4$ from the ETB as the optimization objective and use the higher-level semantic feature $\mathcal{P}_5$ as the auxiliary objective. 

The DRP consists of two branches (as shown in Fig. \ref{Fig.5}), in the first branch, we first expand the channel of auxiliary feature $\mathcal{P}_5$ to match the dimension of the optimization objective and aggregate these feature to obtain feature $\mathcal{I}_{4}$, $i.e.$, $\mathcal{I}_{4}$$=$$\mathcal{C}on(Cat(Ex(\mathcal{P}_5),\mathcal{X}_4))$, where $Ex(\cdot)$ denotes to expand the channel to 128, $\mathcal{C}on(\cdot)$ presents a 1$\times$1 convolution and two 3$\times$3 convolutions. And then feature $\mathcal{I}_{4}$ is separated into the spatial and frequency domains. We perform the Fast Fourier Transform ($fft(\cdot)$) and Inverse Fast Fourier Transform ($ifft(\cdot)$) in the frequency domain and adopt a series of convolution operations ($\mathcal{C}on(\cdot)$) to optimize features in the spatial domain. Subsequently, they are aggregated to obtain the fused feature $\mathcal{N}_{4}^{1}$, that is,
\begin{equation}
\begin{split}
&\mathcal{N}_{4}^{1}=\Phi||ifft(\sigma(fft(\mathcal{I}_4))*fft(\mathcal{I}_4))||+\mathcal{C}on(\mathcal{I}_4),\\ 
\end{split}
\end{equation}
where $\Phi \left \| \cdot \right \|$ denotes the modulus operation. ``$+$'' presents element-wise addition. 
In the second branch, we produce the hybrid reverse attention map ($\mathcal{A}_{r}$, $\mathcal{A}_{r}$$=$$(1-Sig(\mathcal{P}_{5}))+(1-Sig(\Phi||fft(\mathcal{P}_{5})||))$)
) using the auxiliary feature, where $Sig(\cdot)$ denotes the Sigmoid function. Unlike other methods \cite{SINet2,FEDER}, our the reverse attention map ($\mathcal{A}_r$) contains abundant the frequency-spatial information to efficiently obtain the reverse feature $\mathcal{N}_{4}^{2}$, $i.e.$, $\mathcal{N}_{4}^{2}$$=$$\mathcal{A}_{r}*\mathcal{X}_4$. Next, we integrate the features $\mathcal{N}_{4}^{1}$ and $\mathcal{N}_{4}^{2}$ to generate final feature $\mathcal{N}_4$, that is, $\mathcal{N}_4$$=$$C_3Cat(\mathcal{N}_{4}^{1},\mathcal{N}_{4}^{2})$+$\mathcal{P}_{5}$. Subsequently, $\mathcal{N}_{i+1}$ will continue to optimize features $\mathcal{X}_i(i=1,2,3)$  as an auxiliary objective in the proposed DRP. Note that there must be at least one auxiliary feature used for optimizing feature $\mathcal{X}_i$ to generate feature $\mathcal{N}_i$, and auxiliary features are input in the dense connection manner.

\subsection{Loss function}
In the proposed FSEL method, we supervise multi-level feature $\mathcal{N}$ to produce an accurately predicted map. Specifically, we adopt the weighted binary cross-entropy (BCE) and the weighted intersection over union (IoU) \cite{IoU} as the overall loss function to optimize the model based on ground truth ($G$). The loss function can be defined as:
\begin{equation}
	\mathcal{L}_{all}=\sum_{i=1}^{5}\frac{1}{2^{i-1}}  (\mathcal{L}_{bce}^{w}(\mathcal{N}_i,G)+\mathcal{L}_{iou}^{w}(\mathcal{N}_i,G)),
\end{equation}
where $\mathcal{L}_{bce}^{w}$ and $\mathcal{L}_{iou}^{w}$ denote the weighted BCE and IoU functions. $\mathcal{N}_5$ is the feature $\mathcal{P}_5$ from the JDPM.

\begin{table}[t]
\renewcommand{\arraystretch}{0.8}
	\setlength{\tabcolsep}{2pt}
	\centering
        \caption{ Quantitative results on three COD datasets. The best result is shown in \textbf{blod}. ``Ours-R50'', ``Ours-R2N'', and ``Ours-Pvt'' present ResNet50 \cite{ResNet}/Res2Net \cite{R2Net}/PVTv2 \cite{PVT} as backbone. ``Ours-R50$^{\dag}$''denotes using the same input strategy as ZoomNet \cite{ZoomNet}.}
	\resizebox*{1\textwidth}{58mm}{
\begin{tabular}{c|c|cccccc|cccccc|cccccc}
\toprule[1pt]\toprule[1pt]
\multirow{2}{*}{\large Method} & \multirow{2}{*}{Year, Pub.} & \multicolumn{6}{c|}{CAMO (250 images)}        & \multicolumn{6}{c|}{COD10K (2026 images)}     & \multicolumn{6}{c}{NC4K(4121 images)}         \\ \cline{3-20} 
                        &                       &$\mathcal{M}$$\downarrow$   & $F_{\varphi}^{m}$$\uparrow$   & $F_{\varphi}^{a}$$\uparrow$   & $F_{\varphi}^{w}$$\uparrow$   & $S_m$$\uparrow$    & $E_m$$\uparrow$     &$\mathcal{M}$$\downarrow$   & $F_{\varphi}^{m}$$\uparrow$   & $F_{\varphi}^{a}$$\uparrow$   & $F_{\varphi}^{w}$$\uparrow$   & $S_m$$\uparrow$    & $E_m$$\uparrow$     &$\mathcal{M}$$\downarrow$   & $F_{\varphi}^{m}$$\uparrow$   & $F_{\varphi}^{a}$$\uparrow$   & $F_{\varphi}^{w}$$\uparrow$   & $S_m$$\uparrow$    & $E_m$$\uparrow$    \\ \hline\hline
SINet \cite{COD10K}                   & 2020, CVPR             & 0.100 & 0.762 & 0.709 & 0.606 & 0.751 & 0.835 & 0.051 & 0.708 & 0.593 & 0.551 & 0.770 & 0.797 & 0.058 & 0.805 & 0.768 & 0.723 & 0.807 & 0.883 \\ 
UGTR \cite{UGTR}                  & 2021, ICCV             & 0.086 & 0.800 & 0.748 & 0.684 & 0.784 & 0.858 & 0.036 & 0.772 & 0.671 & 0.666 & 0.815 & 0.850 & 0.052 & 0.833 & 0.778 & 0.747 & 0.839 & 0.888 \\ 
JSOCOD \cite{JSOCOD}                 & 2021, CVPR             & 0.073 & 0.812 & 0.779 & 0.728 & 0.800 & 0.872 & 0.035 & 0.762 & 0.705 & 0.684 & 0.807 & 0.882 & 0.047 & 0.838 & 0.803 & 0.771 & 0.841 & 0.906 \\ 
MGL-S \cite{MGL}                   & 2021, CVPR             & 0.089 & 0.791 & 0.733 & 0.664 & 0.772 & 0.850 & 0.037 & 0.765 & 0.667 & 0.655 & 0.808 & 0.851 & 0.055 & 0.826 & 0.771 & 0.731 & 0.828 & 0.885 \\ 
LSR \cite{NC4K}                    & 2021, CVPR             & 0.080 & 0.791 & 0.756 & 0.696 & 0.787 & 0.859 & 0.037 & 0.756 & 0.699 & 0.673 & 0.802 & 0.883 & 0.048 & 0.836 & 0.802 & 0.766 & 0.839 & 0.904 \\ 
PFNet \cite{PFNet}                  & 2021, CVPR             & 0.085 & 0.795 & 0.751 & 0.695 & 0.782 & 0.855 & 0.040 & 0.748 & 0.676 & 0.660 & 0.798 & 0.868 & 0.053 & 0.821 & 0.779 & 0.745 & 0.828 & 0.894 \\ 
SegMaR$_1$ \cite{SegMaR}               & 2022, CVPR             & 0.072 & 0.821 & 0.772 & 0.728 & 0.808 & 0.870 & 0.035 & 0.765 & 0.699 & 0.682 & 0.811 & 0.881 & —     & —     & —     & —     & —     & —     \\ 
PreyNet \cite{PreyNet1}                & 2022, MM               & 0.077 & 0.803 & 0.764 & 0.708 & 0.789 & 0.856 & 0.034 & 0.775 & 0.731 & 0.697 & 0.810 & 0.894 & —     & —     & —     & —     & —     & —     \\ 
FEDER \cite{FEDER}                 & 2023, CVPR             & 0.071 & 0.824 & 0.786 & 0.738 & 0.802 & 0.877 & 0.032 & 0.788 & 0.740 & 0.716 & 0.820 & \textbf{0.901} & 0.044 & 0.852 & \textbf{0.822} & 0.789 & 0.846 & 0.913 \\ \hline
\rowcolor{red!10} Ours-R50                 & —                     & \textbf{0.067} & \textbf{0.833} & \textbf{0.799} & \textbf{0.758} & \textbf{0.821} & \textbf{0.893} & \textbf{0.031} & \textbf{0.802} & \textbf{0.743} & \textbf{0.728} & \textbf{0.830} & 0.898 & \textbf{0.042} & \textbf{0.855} & 0.818 & \textbf{0.792} & \textbf{0.854} & \textbf{0.914} \\ \toprule[1pt]
ZoomNet \cite{ZoomNet}                 & 2022, CVPR            & \textbf{0.066} & 0.832 & 0.792 & 0.752 & 0.820 & 0.883 & \textbf{0.029} & 0.810 & 0.741 & 0.729 & 0.835 & 0.893 & 0.043 & 0.851 & 0.815 & 0.784 & 0.852 & 0.907 \\ \hline
\rowcolor{red!10} Ours-R50$^{\dag}$                  & -            & 0.068 & \textbf{0.837} & \textbf{0.799} & \textbf{0.765} & \textbf{0.826} & \textbf{0.890} & \textbf{0.029} & \textbf{0.814} & \textbf{0.754} & \textbf{0.743} & \textbf{0.839} & \textbf{0.903} & \textbf{0.040} & \textbf{0.863} & \textbf{0.828} & \textbf{0.802} & \textbf{0.861} & \textbf{0.917} \\ \toprule[1pt]
C2FNet \cite{C2FNet}                 & 2021, IJCAI            & 0.080 & 0.803 & 0.764 & 0.719 & 0.796 & 0.865 & 0.036 & 0.764 & 0.703 & 0.686 & 0.811 & 0.886 & 0.049 & 0.832 & 0.788 & 0.762 & 0.838 & 0.901 \\ 
FAPNet \cite{FAPNet}                 & 2022, TIP              & 0.076 & 0.823 & 0.776 & 0.734 & 0.815 & 0.877 & 0.036 & 0.781 & 0.707 & 0.694 & 0.820 & 0.875 & 0.047 & 0.846 & 0.804 & 0.775 & 0.850 & 0.903 \\ 
SINet$_{v2}$ \cite{SINet2}               & 2022, TPAMI            & 0.071 & 0.820 & 0.779 & 0.743 & 0.820 & 0.884 & 0.037 & 0.770 & 0.682 & 0.680 & 0.813 & 0.864 & 0.048 & 0.842 & 0.792 & 0.770 & 0.847 & 0.901 \\ 
BSANet \cite{BSANet}                 & 2022, AAAI             & 0.079 & 0.804 & 0.768 & 0.717 & 0.794 & 0.866 & 0.034 & 0.776 & 0.724 & 0.699 & 0.815 & 0.894 & 0.048 & 0.839 & 0.805 & 0.771 & 0.841 & 0.906 \\ 
BGNet \cite{BGNet}                  & 2022, IJCAI            & 0.073 & 0.825 & 0.786 & 0.749 & 0.811 & 0.878 & 0.033 & 0.795 & \textbf{0.739} & 0.722 & 0.828 & \textbf{0.902} & 0.044 & 0.851 & 0.813 & 0.788 & 0.850 & 0.911 \\ \hline
\rowcolor{red!10} Ours-R2N                 & —                     & \textbf{0.065} & \textbf{0.844} & \textbf{0.803} & \textbf{0.771} & \textbf{0.831} & \textbf{0.895} & \textbf{0.030} & \textbf{0.803} & \textbf{0.739} & \textbf{0.731} & \textbf{0.837} & 0.899 & \textbf{0.042} & \textbf{0.858} & \textbf{0.821} & \textbf{0.798} & \textbf{0.860} & \textbf{0.915} \\ \toprule[1pt]
VST \cite{VST}                    & 2021, ICCV             & 0.081 & 0.812 & 0.753 & 0.713 & 0.808 & 0.853 & 0.037 & 0.779 & 0.721 & 0.698 & 0.817 & 0.882 & 0.048 & 0.840 & 0.801 & 0.768 & 0.844 & 0.899 \\
EVP \cite{EVP}                    & 2023, CVPR             & 0.067 & 0.836 & 0.800 & 0.762 & 0.831 & 0.896 & 0.032 & 0.802 & 0.708 & 0.726 & 0.835 & 0.877 & —     & —     & —     & —     & —     & —     \\ 
FPNet \cite{FPNet}                  & 2023, MM               & 0.056 & 0.863 & 0.838 & 0.802 & 0.851 & 0.912 & 0.029 & 0.817 & 0.765 & 0.755 & 0.847 & 0.909 & —     & —     & —     & —     & —     & —     \\ 
HiNet \cite{HitNet}                  & 2023, AAAI             & 0.055 & 0.857 & 0.833 & 0.809 & 0.849 & 0.910 & 0.023 & 0.850 & \textbf{0.818} & \textbf{0.806} & 0.868 & \textbf{0.936} & 0.037 & 0.879 & 0.854 & 0.834 & 0.874 & 0.928 \\ 
FSPNet \cite{FSPNet}                 & 2023, CVPR             & 0.050 & 0.869 & 0.829 & 0.799 & 0.855 & 0.919 & 0.026 & 0.816 & 0.736 & 0.735 & 0.847 & 0.900 & 0.035 & 0.878 & 0.826 & 0.816 & 0.878 & 0.923 \\ 
SAM \cite{SAM}                  & 2023, ICCV               & —     & —     & —     & —     & —     & — & 0.050     & 0.844     & 0.758     & 0.701     & 0.778     & 0.800 & 0.078     & 0.852     & 0.754     & 0.696     & 0.765     & 0.778     \\ \hline
\rowcolor{red!10}Ours-Pvt                 & —                     & \textbf{0.040} & \textbf{0.891} & \textbf{0.864} & \textbf{0.851} & \textbf{0.885} & \textbf{0.942} & \textbf{0.021} & \textbf{0.853} & 0.796 & 0.800 & \textbf{0.873} & 0.928 & \textbf{0.030} & \textbf{0.895} & \textbf{0.864} & \textbf{0.853} & \textbf{0.892} & \textbf{0.941} \\ \toprule[1pt]\toprule[1pt]
\end{tabular}}
\label{Table.1}
\end{table}
\begin{table}[t]
\renewcommand{\arraystretch}{0.9}
	\setlength{\tabcolsep}{2pt}
	\centering
        \caption{Efficiency analysis of our FSEL and multiple COD methods.}
	\resizebox*{1\textwidth}{7mm}{
\begin{tabular}{c|cccccccccccccc}
\toprule[1pt]
         & SINet \cite{COD10K} & PFNet\cite{PFNet} & MGL\cite{MGL}    & UGTR\cite{UGTR}    & JSOCOD\cite{JSOCOD} & C2FNet\cite{C2FNet} & ZoomNet\cite{ZoomNet} & SegMaR\cite{SegMaR} & FSPNet\cite{FSPNet} & HitNet\cite{HitNet} & FEDER\cite{FEDER} & Ours-R50 & Ours-R2N & Ours-Pvt \\ \hline
Parameters (M) & 48.95 & 46.50 & 63.60  & 48.87   & 217.98 & 28.41  & 32.38   & 55.62  & 273.79 & \textbf{25.73}  & 37.37 & 29.15  & 29.31   & 67.13  \\ \hline
FLOPs (G)    & 38.75 & 53.22 & 553.94 & 1000.01 & 112.34 & 26.17  & 203.50  & 33.65  & 283.31 & 56.55  & \textbf{23.98} & 35.64  & 37.07   & 54.73  \\ \toprule[1pt]
\end{tabular}}
\label{table2}
\end{table}

\section{Experiment}
\subsection{Experimental Setups}
\textbf{Datasets.} We evaluate our FSEL model on three benchmark datasets: CAMO \cite{CAMO}, COD10K \cite{COD10K}, and NC4K \cite{NC4K}. CAMO \cite{CAMO} is an early dataset that contains 1,250 camouflaged images with 1,000 training images and 250 testing images. COD10K \cite{COD10K} is a currently large dataset of camouflaged objects, consisting of 3,040 training images and 2,026 testing images. NC4K \cite{NC4K} is the largest COD dataset for testing, containing 4,121 images of camouflaged objects. We use 4,040 images from CAMO \cite{CAMO} and COD10K \cite{COD10K} as training samples to train the FSEL.

\textbf{Implementation details.} The proposed FSEL model is implemented in the PyTorch framework on four NVIDIA GTX 4090 GPUs with 24GB. We utilize the pre-trained PVTv2 \cite{PVT}/ResNet50 \cite{ResNet}/Res2Net \cite{R2Net} as the encoder to extract initial features. Following \cite{SINet2,FEDER}, we also employ data augmentation techniques such as random flipping and random clipping to enhance training data. We use the Adam optimizer with an initial learning rate of 1e-4 and decay the rates by 10 every 60 epochs. All input images are resized to 416$\times$416, and the batch size is set to 40 for 180 epochs of training progressing.

\textbf{Evaluation metrics.} We use six well-known evaluation metrics, including Mean Absolute Error ($\mathcal{M}$), Maximum F-measure ($F_{\varphi}^{m}$), Average F-measure ($F_{\varphi}^{a}$), Weighted F-measure ($F_{\varphi}^{w}$), S-measure ($S_m$), and E-measure ($E_m$). 

\begin{figure*}[t]
	\centering\includegraphics[width=0.83\textwidth,height=3.2cm]{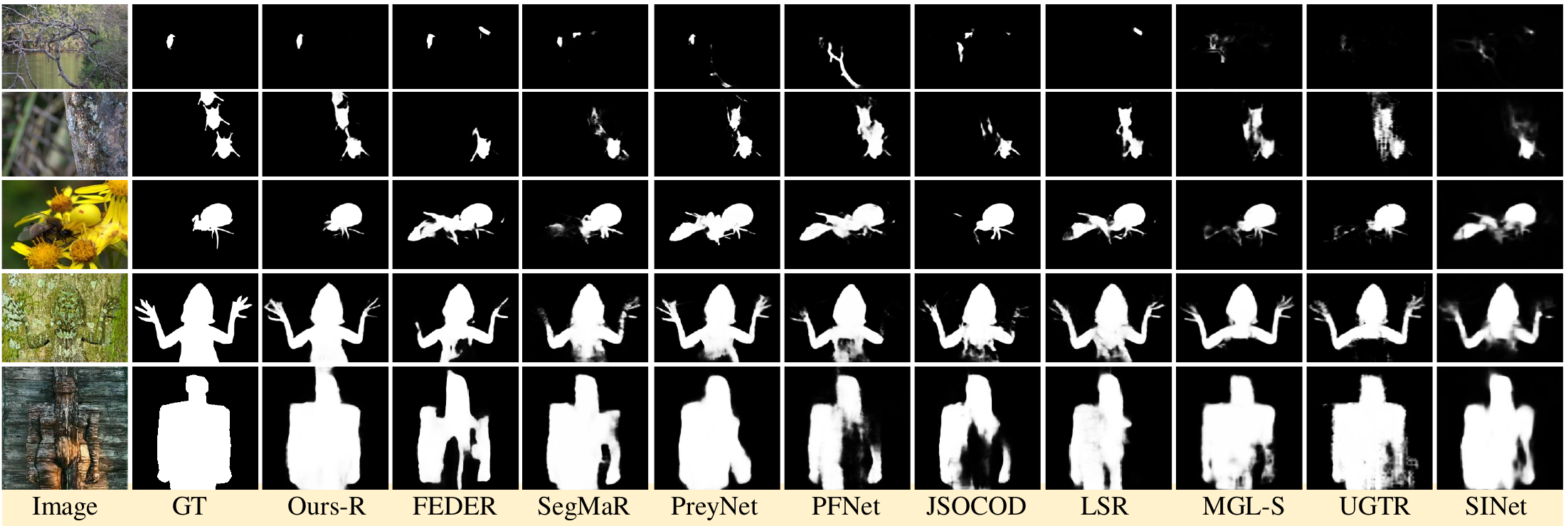}
	\caption{Qualitative comparisons of the proposed FSEL and nine COD methods.}
	\label{Fig.6}
\end{figure*}
\subsection{Comparisons with the SOTAs}

We conduct a comparison of our FSEL with twenty-one COD methods, including SINet \cite{COD10K}, C2FNet \cite{C2FNet}, UGTR \cite{UGTR}, JSOCOD \cite{JSOCOD}, MGL-S \cite{MGL}, LSR \cite{NC4K}, PFNet \cite{PFNet}, VST \cite{VST}, FAPNet \cite{FAPNet}, SINet$_{v2}$ \cite{SINet2}, BSANet \cite{BSANet}, SegMaR \cite{SegMaR}, ZoomNet \cite{ZoomNet}, BGNet \cite{BGNet}, PreyNet \cite{PreyNet1}, FEDER \cite{FEDER}, EVP \cite{EVP}, FPNet \cite{FPNet}, HitNet \cite{HitNet}, FSPNet \cite{FSPNet}, and SAM \cite{SAM}. Note that the predicted maps from all methods are provided by the authors or obtained from open-source codes.

\textbf{Quantitative Evaluation.} Table \ref{Table.1} summarizes the quantitative result of our FSEL and other 21 SOTA models. From Table \ref{Table.1}, we can observe that the FSEL model achieves excellent performance across different backbone networks. Particularly, compared to the recently proposed FEDER \cite{FEDER} method, our FSEL with ResNet50 \cite{ResNet} backbone overall surpasses 5.97\%, 3.23\%, and 4.76\% on three public datasets under the $\mathcal{M}$ metric. Besides, in the Res2Net \cite{R2Net} backbone, our FSEL achieves average performance gains of 9.23\%, 2.30\%, 2.16\%, 2.94\%, 1.34\%, and 1.24\% over the second-best method in terms of six public evaluation metrics on CAMO \cite{CAMO} dataset. Moreover, compared to the frequency-based FPNet \cite{FPNet} and EVP \cite{EVP} methods, FSEL method with PVTv2 \cite{PVT} backbone achieves average performance gains of 38.10\%, 4.41\%, 4.05\%, 5.96\%, 3.07\%, and 2.09\% over FPNet \cite{FPNet} and 52.38\%, 6.36\%, 12.43\%, 10.19\%, 4.55\%, and 5.82\% over EVP \cite{EVP} in terms of $\mathcal{M}$, $F_{\varphi}^{m}$, $F_{\varphi}^{a}$, $F_{\varphi}^{w}$, $S_m$, and $E_m$ on the COD10K \cite{COD10K} dataset. Furthermore, FSEL achieves excellent performance when adopting the same input strategy with ZoomNet \cite{ZoomNet}. The superiority in performance benefits from the joint optimization of the ETB, JDPM, and DRP for input features in the frequency and spatial domains. In addition, we provide the parameters and FLOPs in Table \ref{table2}. It can be seen that the proposed FSEL method parameters and FLOPs are at a medium to high level, however, our performance far exceeds that of methods with similar parameters and FLOPs.

\textbf{Qualitative Evalation.} Fig. \ref{Fig.6} gives the visual comparisons between our FSEL and several COD method in different scenarios. As depicted in Fig. \ref{Fig.6}, the proposed FSEL method exhibits accurate and complete segmentation for camouflaged objects with different sizes compared to current COD methods ($i.e.$, HitNet \cite{HitNet}, FSPNet \cite{FSPNet}, and FPNet \cite{FPNet}). These visual results demonstrate the superiority of the FSEL method for detecting camouflaged objects through the frequency-spatial domain optimization strategy.
\begin{table}[t]
\renewcommand{\arraystretch}{0.9}
	\setlength{\tabcolsep}{2pt}
	\centering
        \caption{Ablation analysis of our FSEL structure.}
	\resizebox{1\textwidth}{16mm}{
\begin{tabular}{c|cccc|cccccc|cccccc}
\toprule[1pt]\toprule[1pt]
\multicolumn{1}{l|}{\multirow{2}{*}{No.}} & \multicolumn{4}{c|}{Structure Setting} & \multicolumn{6}{c|}{CAMO(250 images)}         & \multicolumn{6}{c}{COD10K(2026 images)}       \\ \cline{2-17} 
\multicolumn{1}{l|}{}                     & Baseline    & ETB    & DRP    & JDPM   &$\mathcal{M}$$\downarrow$   & $F_{\varphi}^{m}$$\uparrow$   & $F_{\varphi}^{a}$$\uparrow$   & $F_{\varphi}^{w}$$\uparrow$   & $S_m$$\uparrow$    & $E_m$$\uparrow$    &$\mathcal{M}$$\downarrow$   & $F_{\varphi}^{m}$$\uparrow$   & $F_{\varphi}^{a}$$\uparrow$   & $F_{\varphi}^{w}$$\uparrow$   & $S_m$$\uparrow$    & $E_m$$\uparrow$    \\ \hline\hline
\textbf{\color{red}(a)}                                         & $\checkmark$           &        &        &        & 0.093 & 0.784 & 0.712 & 0.663 & 0.767 & 0.847 & 0.046 & 0.749 & 0.617 & 0.610 & 0.778 & 0.818 \\ 
\textbf{\color{red}(b)}                                         & $\checkmark$           & $\checkmark$      &        &        & 0.076 & 0.811 & 0.763 & 0.723 & 0.801 & 0.873 & 0.034 & 0.789 & 0.712 & 0.702 & 0.821 & 0.886 \\ 
\textbf{\color{red}(c)}                                         & $\checkmark$           &        & $\checkmark$      &        & 0.074 & 0.820 & 0.778 & 0.735 & 0.810 & 0.876 & 0.034 & 0.794 & 0.722 & 0.713 & 0.826 & 0.887 \\
\textbf{\color{red}(d)}                                         & $\checkmark$           &        &       & $\checkmark$       & 0.081 & 0.797 & 0.743 & 0.697 & 0.787 & 0.863 & 0.039 & 0.769 & 0.676 & 0.668 & 0.804 & 0.863 \\
\textbf{\color{red}(e)}                                         & $\checkmark$           & $\checkmark$      & $\checkmark$       &     & 0.074 & 0.823 & 0.782 & 0.731 & 0.807 & 0.875 & 0.032 & 0.798 & 0.723 & 0.714 & 0.828 & 0.888 \\ 
\textbf{\color{red}(f)}                                         & $\checkmark$           & $\checkmark$       &      & $\checkmark$      & 0.071 & 0.807 & 0.767 & 0.728 & 0.802 & 0.878 & 0.033 & 0.781 & 0.719 & 0.702 & 0.815 & 0.895 \\ 
\textbf{\color{red}(g)}                                         & $\checkmark$           &       & $\checkmark$     & $\checkmark$      & 0.071 & 0.830 & 0.789 & 0.742 & 0.810 & 0.879 & \textbf{0.031} & 0.796 & 0.730 & 0.718 & 0.827 & 0.893 \\ \hline
\rowcolor{red!10}\textbf{\color{red}(h)}                                        & $\checkmark$           & $\checkmark$      & $\checkmark$      & $\checkmark$      & \textbf{0.067} & \textbf{0.833} & \textbf{0.799} & \textbf{0.758} & \textbf{0.821} & \textbf{0.893} & \textbf{0.031} & \textbf{0.802} & \textbf{0.743} & \textbf{0.728} & \textbf{0.830} & \textbf{0.898} \\ \toprule[1pt]\toprule[1pt]
\end{tabular}}
\label{table3}
\end{table}

\subsection{Ablation Study}
\begin{figure}[t]
		\centering\includegraphics[width=0.75\textwidth,height=3.3cm]{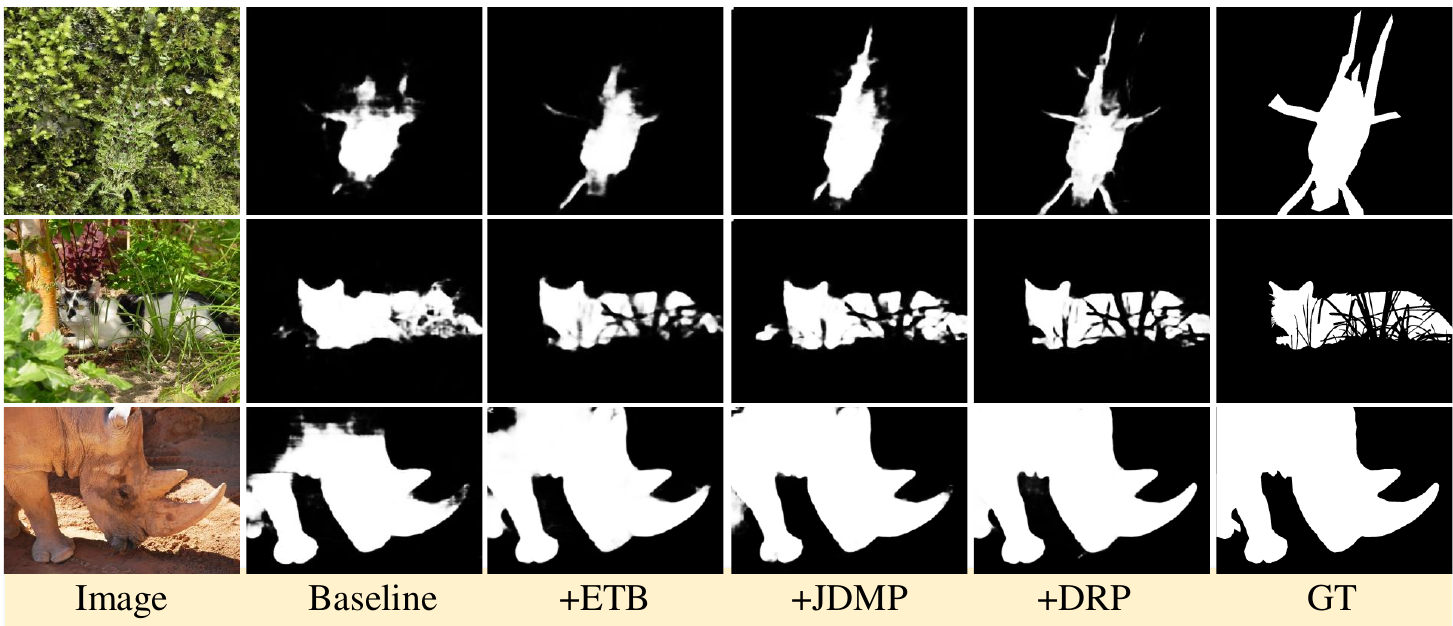}
		\caption{ Visual results of the effectiveness of our modules.}
		\label{Fig.7}
\end{figure}

\textbf{Effectiveness of proposed each component.} We provide the quantitative results of different components in the proposed FSEL model, shown in Table \ref{table3}. Specifically, we first adopt ``ResNet50 \cite{ResNet} - FPN \cite{FPN}'' as ``Baseline'' (Tab. \ref{table3}{\color{red}(a)}) to detect camouflaged objects. And then we independently validate the effectiveness of ``ETB'' (Tab. \ref{table3}{\color{red}(b)}), ``DRP'' (Tab. \ref{table3}{\color{red}(c)}) and ``JDPM'' (Tab. \ref{table3}{\color{red}(d)}), and it can be seen that the performance of the predicted map increases significantly when the proposed component is embedded in the ``Baseline'' (Tab. \ref{table3}{\color{red}(a)}). Additionally, we validate the compatibility among all modules. From Tab. \ref{table3}{\color{red}(e)}, Tab. \ref{table3}{\color{red}(f)}, and Tab. \ref{table3}{\color{red}(g)}, it can be observed that the three components are compatible with each other. Subsequently, all components are integrated, and the performance of the model is improved once again, as shown in Tab. \ref{table3}{\color{red}(h)}. Additionally, in Fig. \ref{Fig.7}, we show the visual results obtained by progressively adding the proposed components ($i.e.$, ETB, JDPM, and DRP), generating that the predicted map gradually approaches the ground truth (GT). The above results demonstrate the effectiveness of our proposed modules in detecting camouflaged objects.

\textbf{Effectiveness of frequency-spatial information within the ETB.} Do we really need frequency information? To answer this question, we perform a series of experiments in the internal part of the ETB. Specifically, the ETB is first divided into two parts, with ``ETB-S'' (Tab. \ref{table4}{\color{red}(a)}) containing only spatial information, and ``ETB-F'' (Tab. \ref{table4}{\color{red}(b)}) presenting that it includes only frequency information. Based on Table \ref{table4}, the performance of the separate frequency and spatial domains exhibits certain differences compared to the complete ETB (Tab. \ref{table4}{\color{red}(g)}). Besides, we investigate the entanglement learning of frequency-spatial information in the proposed ETB. In Tab. \ref{table4}{\color{red}(c)}-{\color{red}(f)}, it can be seen that the frequency and spatial features interact fusion to achieve entanglement between two states, enhancing the model's reasoning ability of camouflaged objects.
\begin{table}[t]
\renewcommand{\arraystretch}{0.9}
\setlength{\tabcolsep}{2pt}
\centering
\caption{Ablation analysis within the ETB structure.}
\resizebox{1\textwidth}{15mm}{
\begin{tabular}{c|cccc|cccccc|cccccc}
\toprule[1pt]\toprule[1pt]
\multicolumn{1}{l|}{\multirow{2}{*}{No.}} & \multicolumn{4}{c|}{ETB} & \multicolumn{6}{c|}{CAMO(250 images)}         & \multicolumn{6}{c}{COD10K(2026 images)}       \\ \cline{2-17} 
\multicolumn{1}{l|}{}                     & FSA  & SFA  & FFFN  & SFFA & $\mathcal{M}$$\downarrow$   & $F_{\varphi}^{m}$$\uparrow$   & $F_{\varphi}^{a}$$\uparrow$   & $F_{\varphi}^{w}$$\uparrow$   & $S_m$$\uparrow$    & $E_m$$\uparrow$    &$\mathcal{M}$$\downarrow$   & $F_{\varphi}^{m}$$\uparrow$   & $F_{\varphi}^{a}$$\uparrow$   & $F_{\varphi}^{w}$$\uparrow$   & $S_m$$\uparrow$    & $E_m$$\uparrow$    \\ \hline

\textbf{\color{red}(a)}                                         &       & $\checkmark$     &        & $\checkmark$     & 0.079 & 0.798 & 0.752 & 0.706 & \textbf{0.802} & 0.864 & 0.037 & 0.770 & 0.697 & 0.679 & 0.816 & 0.874 \\ 
\textbf{\color{red}(b)}                                         & $\checkmark$     &       & $\checkmark$      &       & \textbf{0.075} & 0.810 & \textbf{0.770} & \textbf{0.725} & 0.798 & \textbf{0.880} & \textbf{0.034} & 0.778 & \textbf{0.713} & 0.695 & 0.812 & \textbf{0.890} \\ 
\textbf{\color{red}(c)}                                         & $\checkmark$     &       & $\checkmark$      & $\checkmark$     & \textbf{0.075} & 0.807 & 0.759 & 0.720 & 0.795 & 0.876 & 0.035 & 0.778 & 0.692 & 0.687 & 0.811 & 0.873 \\ 
\textbf{\color{red}(d)}                                         &       & $\checkmark$     & $\checkmark$      & $\checkmark$     & 0.078 & \textbf{0.811} & 0.764 & 0.723 & 0.800 & 0.875 & \textbf{0.034} & 0.777 & 0.703 & 0.692 & 0.814 & 0.882 \\ 
\textbf{\color{red}(e)}                                         & $\checkmark$     & $\checkmark$     & $\checkmark$      &       & 0.076 & 0.808 & 0.764 & 0.712 & 0.795 & 0.872 & \textbf{0.034} & 0.782 & 0.702 & 0.692 & 0.819 & 0.878 \\ 
\textbf{\color{red}(f)}                                         & $\checkmark$     & $\checkmark$     &        & $\checkmark$     & 0.077 & 0.801 & 0.757 & 0.712 & 0.794 & 0.873 & 0.036 & 0.778 & 0.697 & 0.686 & 0.814 & 0.875 \\ \hline
\rowcolor{red!10}\textbf{\color{red}(g)}                                         & $\checkmark$     & $\checkmark$     & $\checkmark$      & $\checkmark$     & 0.076 & \textbf{0.811} & 0.763 & 0.723 & 0.801 & 0.873 & \textbf{0.034} & \textbf{0.789} & 0.712 & \textbf{0.702} & \textbf{0.821} & 0.886 \\ \toprule[1pt]\toprule[1pt]
\end{tabular}}
\label{table4}
\end{table}

\subsection{Expanded application}
\begin{figure}[t]
		\centering\includegraphics[width=0.70\textwidth,height=3.3cm]{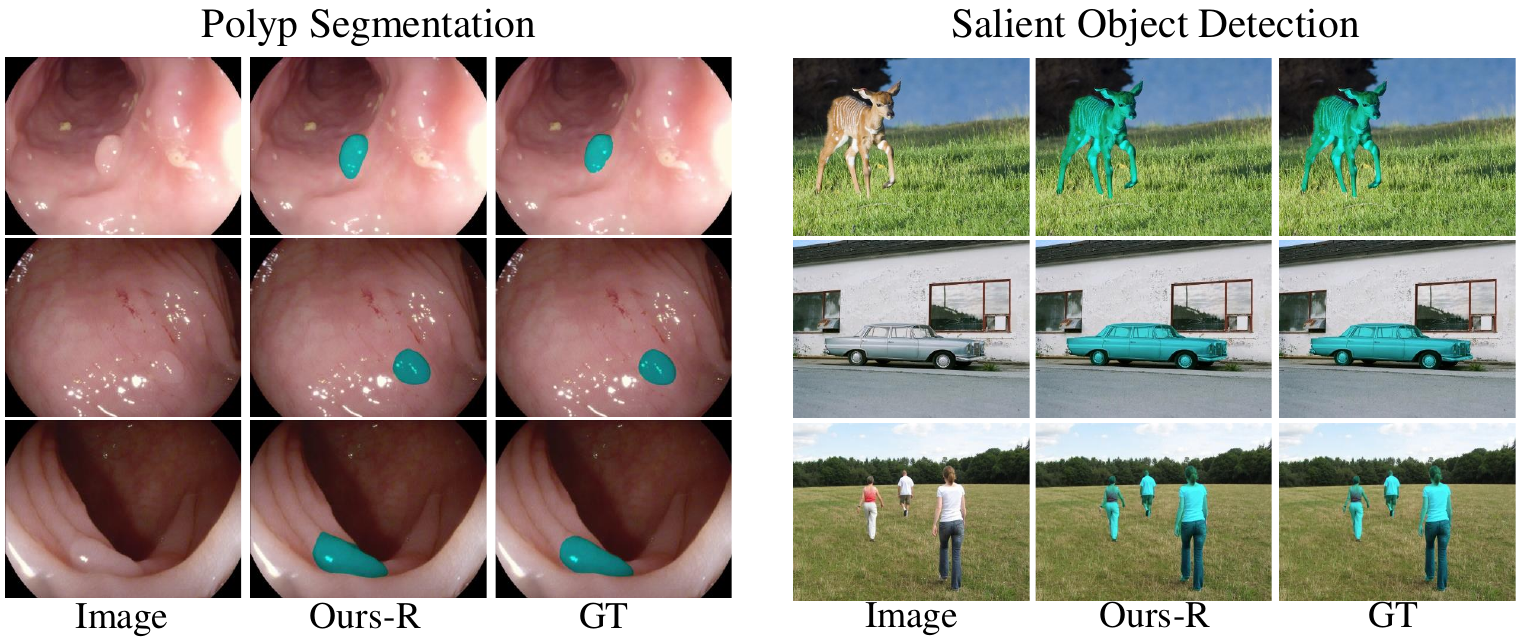}
		\caption{ Visual results of the expanded application.}
		\label{Fig.8}
\end{figure}

To demonstrate the generalization ability of our FSEL model, we extend the FSEL model to salient object detection and polyp segmentation tasks. As shown in Fig. \ref{Fig.8}, the proposed FSEL method achieves highly accurate segmentation for both salient objects and polyps, benefiting from the complementary utilization of frequency domain and spatial information. More details and data are presented in the \textbf{supplementary materials}.

\section{Conclusion}

In this paper, we introduce a new approach for detecting camouflaged objects called Frequency-Spatial Entanglement Learning (FSEL). The key to FSEL is to extract important information from both the frequency and spatial domains. To achieve this, we have developed a Joint Domain Perception Module that combines multi-scale information from frequency-spatial features to accurately localize regions. Additionally, we have created an Entanglement Transformer Block that can be easily integrated into existing methods to improve their performance by modeling long-range dependencies in the hybrid domain. Furthermore, we have designed a Dual-Domain Reverse Parser that interacts with diverse information in multi-layer features to achieve more precise segmentation. Our extensive comparison experiments demonstrate that FSEL outperforms 21 state-of-the-art COD methods on three popular benchmark datasets.

\subsubsection{Acknowledgments.} This work was supported in part by the National Science Fund of China (No. 62276135, 62361166670, 62372238, and 62302006).
%
%
\bibliographystyle{splncs04}
\bibliography{egbib}
\end{document}